%% file: 0-main.tex
\documentclass[letterpaper]{article} 
\usepackage{aaai23}  
\usepackage{times}  
\usepackage{helvet}  
\usepackage{courier}  
\usepackage[hyphens]{url}  
\usepackage{graphicx} 
\usepackage{natbib}  
\usepackage{caption} 
\usepackage{algorithm}
\usepackage{algorithmic}

\usepackage{newfloat}
\usepackage{listings}
\DeclareCaptionStyle{ruled}{labelfont=normalfont,labelsep=colon,strut=off} 
\floatstyle{ruled}
\newfloat{listing}{tb}{lst}{}
\floatname{listing}{Listing}
\title{Explaining Model Confidence Using Counterfactuals}
\author {
    Thao Le, Tim Miller, Ronal Singh, Liz Sonenberg
}
\affiliations {
    School of Computing and Information Systems, 
    The University of Melbourne\\
    thaol4@student.unimelb.edu.au,
    \{tmiller, rr.singh, l.sonenberg\}@unimelb.edu.au
}

\usepackage{bibentry}

\usepackage{graphicx}
\usepackage{amsmath}
\usepackage{booktabs}
\usepackage{verbatim}

\DeclareMathOperator*{\argmin}{arg\,min}

\usepackage{subcaption}
\usepackage{bm}
\usepackage{multirow}
\usepackage{algorithm}
\usepackage{algorithmic}
\usepackage{placeins}
\usepackage{tikz}
\newcommand*{\checkmark}[1][]{\tikz[x=1em, y=1em]\fill[#1] (0,.35) -- (.25,0) -- (1,.7) -- (.25,.15) -- cycle;}
\usepackage{multirow}
\usepackage{dcolumn}
\newcolumntype{C}{D{(}{(}{5}}
\usepackage[skip=6pt]{caption}
\usepackage[disable]{todonotes}
\newcommand{\liz}[2][]{\todo[color=cyan!80, #1, inline]{Liz: {\small #2}}}

\begin{document}

\maketitle

\begin{abstract}
Displaying confidence scores in human-AI interaction has been shown to help build trust between humans and AI systems. However, most existing research uses only the confidence score as a form of communication. As confidence scores are just another model output, users may want to understand why the algorithm is confident to determine whether to accept the confidence score. In this paper, we show that counterfactual explanations of confidence scores help study participants to better understand and better trust a machine learning model's prediction. We present two methods for understanding model confidence using counterfactual explanation: (1) based on counterfactual examples; and (2) based on visualisation of the counterfactual space. Both increase understanding and trust for study participants over a baseline of no explanation, but qualitative results show that they are used quite differently, leading to recommendations of when to use each one and directions of designing better explanations.
\end{abstract}

\input{1-introduction}
\input{2-background}
\input{3-model}
\input{4-experiment}
\input{6-conclusion}

\section{Acknowledgments}
This research was supported by the University of Melbourne Research Scholarship (MRS) and by Australian Research Council (ARC) Discovery Grant DP190103414: Explanation in Artificial Intelligence: A Human-Centred Approach.

\bibliography{references}

\end{document}

%% file: 1-introduction.tex
\section{Introduction}

Explaining why an AI model gives a certain prediction can promote trust and understanding for users, especially for non-expert users. While recent research~\cite{Zhang20effect,Wang21uncertainty} has used confidence (or uncertainty) measures as a way to improve AI model understanding and trust, the area of explaining why the AI model is confident (or not confident) in its prediction is still underexplored~\cite{TomsettPBCCSPK20}.

In Machine Learning (ML), the \textit{confidence score} indicates the chances that the ML model's prediction is correct. In other words, it shows how \textit{certain} the model is in its prediction, which can be defined as the predicted probability for the best outcome~\cite{Zhang20effect}. Another way to define the \textit{confidence score} is based on uncertainty measures, which can be calculated using entropy ~\cite{Bhatt20uncertainty} or using \textit{uncertainty sampling}~\cite{LewisG94},~\cite[p93]{monarch2021human}.

In this paper, we complement prior research by applying a counterfactual (CF) explanation method to generate explanations of the confidence of a predicted output. It is increasingly accepted that explainability techniques should be built on research in philosophy, psychology and cognitive science~\cite{Miller19social,Byrne19} and that the evaluation process of explanation should involve human-subject studies~\cite{Miller17Beware,ForsterHKK21,KennyFQK21,WaaNCN21}. We therefore evaluate our explanation to know whether counterfactual explanations can improve \textit{understanding}, \textit{trust}, and \textit{user satisfaction} in two user studies using existing methods for assessing understanding, trust and satisfaction. We present the CF explanation using two designs: (1) providing counterfactual examples (example-based counterfactuals); and (2) visualising the counterfactual space for each feature and its effect on model confidence (visualisation-based counterfactuals).

Our contributions are:
\begin{itemize}
    \item We formalise two approaches for the counterfactual explanation of confidence score: one using counterfactual examples and one visualising the counterfactual space.  
    \item Through two user studies we demonstrate that showing counterfactual explanations of confidence scores can help users better understand and trust the model.
    \item Using qualitative analysis, we observe limits of the  two explainability approaches and suggest directions for improving presentations of counterfactual explanations.
\end{itemize}

%% file: 2-background.tex
\section{Background and Related Work}
In this section, we review related work on counterfactual explanations and confidence (or uncertainty) measures.

\subsection{Counterfactual Explanations}
Counterfactual explanation is described as the possible smallest changes in input values in order to change the model prediction to a desired output~\cite{wachter2017counterfactual}. It has been increasingly used in explainable AI (XAI) to facilitate human interaction with the AI model~\cite{Miller19social,Miller2021-ql,Byrne19,ForsterHKK21}. Counterfactual explanations can be expressed in the following example: ``You were denied a loan because your annual income was \$30,000. If your income had been \$45,000, you would have been offered a loan". To generate counterfactuals, \cite{wachter2017counterfactual} suggest finding solutions of the following loss function.
\begin{equation}
\argmin_{x'}\max_{\lambda} \lambda(f(x') - y')^2 + d(x,x')
\label{eq:loss-wachter}	
\end{equation}
where $x'$ is the counterfactual solution; $(f(x') - y')^2$ presents the distance between the model's prediction output of counterfactual input $x'$ and the desired counterfactual output $y'$; $d(x,x')$ is the distance between the original input and the counterfactual input; and $\lambda$ is a weight parameter. A high $\lambda$ means we prefer to find counterfactual point $x'$ that gives output $f(x')$ close to the desired output $y'$, a low $\lambda$ means we aim to find counterfactual input $x'$ that is close to the original input $x$ even when the counterfactual output $f(x')$ can be far away from the desired output $y'$. In this model, $f(x)$ would be the predicted output, such as a denied loan, and $y'$ would be the desired output -- the loan is granted. The counterfactual $x'$ would be the properties of a similar customer that would have received the loan. Equation~\ref{eq:loss-wachter} can be solved by using the Lagrangian approach. However, this approach has stability issues~\cite{Russell19}. Therefore, \citeauthor{Russell19} \shortcite{Russell19} proposes another search algorithm to generate counterfactual explanations based on mixed-integer programming, assumed where input variables can be continuous or categorical values. They defined a set of linear integer constraints, which is called \textit{mixed polytope}. These constraints can be given to Gurobi Optimization~\cite{gurobi_2022} and then an optimal solution is generated.

\citeauthor{Antoran21explainUncertainty} \shortcite{Antoran21explainUncertainty} propose Counterfactual Latent Uncertainty Explanations (CLUE), to identify features  responsible for the model's uncertainty. Their idea for showing counterfactual examples is similar to ours, however we go further by considering ways to visualise the counterfactual space, run a more comprehensive user study to measure understanding, satisfaction, and trust, and undertake a qualitative analysis to identify limitations of current approaches.
 
There are many other approaches to solving counterfactuals for tabular~\cite{MothilalST20,KeaneS20}, image~\cite{GoyalWEBPL19,dhurandhar-contrastive}, text~\cite{JacoviSRECG21,RiveiroT21} and time series data~\cite{DelaneyGK21}. None of these are for explaining model confidence, however, the underlying algorithms could be modified to search over the model confidence instead of the model output. 

\subsection{Confidence (Uncertainty) Measures}

A confidence score measures how confident a ML model is in its prediction; or inversely, how uncertain it is. A common method of measuring uncertainty is to use the prediction probability~\cite{Delaney21,Bhatt20uncertainty}. Specifically, \textit{uncertainty sampling }~\cite{LewisG94} is an approach that queries unlabelled instance $x$ with maximum uncertainty to get human feedback. There are four types of uncertainty sampling~\cite[p70]{monarch2021human}: \textit{Least confidence}, \textit{Margin of confidence}, \textit{Ratio of confidence} and \textit{Entropy}. \citeauthor{Zhang20effect} \shortcite{Zhang20effect} demonstrate that communicating confidence scores can support trust calibration for end users. \citeauthor{Wang21uncertainty} \shortcite{Wang21uncertainty} also argue that showing feature attribution uncertainty helps improve model understanding and trust.

\citeauthor{WaaSDN20} \shortcite{WaaSDN20} propose a framework called \textit{Interpretable Confidence Measures (ICM)} which provides predictable and explainable confidence measures based on case-based reasoning~\cite{Atkeson97}. Case-based reasoning provides prediction based on similar past cases of the current instance. This approach however did not address counterfactual explanations of model confidence.

%% file: 3-model.tex
\section{Formalising Counterfactual Explanation of Confidence}

This section describes two methods for CF explanation: one based on counterfactual examples~\cite{Antoran21explainUncertainty} and one based on counterfactual visualisation as in Figure~\ref{fig:graph-explanation-discrete}.

\subsection{Generating Counterfactual Explanation of Confidence}

In this section, we show how to generate counterfactual explanations of the confidence score in data where input variables can take either categorical or continuous values. The counterfactual model can generate explanations to either increase or decrease the confidence score of a specific class. For example, when the AI model predicts that an employee will leave the company with confidence of $70\%$, a person may ask: \emph{Why is the model 70\% confident instead of 40\% confident or less?}. This person could ask why the model prediction did not have a lower confidence score when they were sceptical about the high confidence score. We aim to generate counterfactual inputs that bring the confidence score to 40\% or lower. An example of counterfactual explanation in this case is: \textit{``One way you could have got a confidence score of 40\% instead is if Daily Rate had taken the value 400 rather than 300"}. Therefore, from this counterfactual explanation, we know that we can achieve lowering of the confidence of them resigning from the company by increasing the employee's daily rate.

We now describe our approach to generate counterfactuals for confidence scores. We follow \citeauthor{Russell19} \shortcite{Russell19} in proposing an algorithm to search for counterfactual points of output confidence. Importantly, we modify this approach to find counterfactual points that change the confidence score but do not change the predicted class. 

Formally, given a question: ``Why does the model prediction have a confidence score of $U(x)$ rather than greater than (or less than) $T$?" where $T$ is a user-defined confidence threshold, $x$ is the input instance, $U(x)$ is the confidence score of the original prediction, we want to find the counterfactual explanation of confidence $U(x')$  generated by data point $x'$ such that $U(x') > T$ or $U(x') < T$ depending on the question. In case the user cannot give a threshold T, the default threshold T value is the original confidence score $U(x)$ of the prediction. We seek the counterfactual point $x'$ by solving Equation~\ref{eq:search-counterfactual}: 

\begin{equation}
    \argmin_{x'} ( ||x - x'||_{1,w} + |U(x') - T| )
    \label{eq:search-counterfactual}
\end{equation}

such that:
\begin{align}
	U(x') > T && \text{if } T > U(x)  \label{eq:contrastive-greater} \\
	U(x') < T && \text{if } T < U(x) \label{eq:contrastive-lesser}
\end{align}
\begin{equation}
    \begin{cases}
        P(y=k \mid x') < D & \text{if } P(y=k \mid x) < D \\
        P(y=k \mid x') \geq D & \text{if } P(y=k \mid x) \geq D
    \end{cases}
    \label{eq:same-space}
\end{equation}

where $||.||_{1,w}$ is a weighted $l_1$ norm with weight $w$ defined as the inverse median absolute deviation (MAD)~\cite{wachter2017counterfactual}; $D$ is the decision boundary that classifies the class.

We apply Equation~\ref{eq:contrastive-greater} when we want to find counterfactual $x'$ that increases the confidence score, and Equation~\ref{eq:contrastive-lesser} for a counterfactual $x'$ that decreases the confidence score. Since $x$ and $x'$ will give the same output prediction as class $k$ but different confidence scores $U(x)$ and $U(x')$, $P(x)$ and $P(x')$ must be in the same space according to the decision boundary, defined as Equation~\ref{eq:same-space}.

\subsection{Example-Based Counterfactual Explanation}

Given the original instance input shown in column \textit{Original Value} in Table~\ref{tab:table-explanation}, the AI model predicts that this person has an income of \textit{Lower than $\$50,000$} with a confidence score of $57.8\%$. We choose a factual confidence score $T = 45\%$ and search for $x'$ where $U(x') < T$. An example of counterfactual explanation generated using our method is:
\textit{``One way you could have got a confidence score of less than $45\%$ ($30.1\%$) instead is if Occupation had taken value Manager rather than Service."}

\begin{table}[t]
	\centering \small
	\setlength{\tabcolsep}{2pt} 
  \begin{tabular}{lccc}
    \toprule
  \textbf{Attribute} & \textbf{Alternative 1}  & \textbf{Alternative 2} & \textbf{Original}\\
  \midrule
Marital status & - & - & Married\\
Years of education&	-&	- & 9\\
\textbf{Occupation} & \textbf{Manager} & \textbf{Skilled} & \textbf{Service}\\
& & \textbf{Specialty} & \\
Age & - & - & 63\\
Any capital gains&	-&	- & No\\
Working hours &-&-& 12\\
per week  & & & \\
Education &	-  & - & High School \\
\midrule
\textbf{Confidence score} &	\bm{$30.1\%$} & \bm{$42.1\%$} &	\bm{$57.8\%$}\\
\midrule
\textbf{AI prediction}	& \multicolumn{3}{c}{\textbf{Lower than \$50,000}} \\
  \bottomrule
\end{tabular}
\caption{Example-based counterfactual explanation presented in a table. In alternative columns, notation (-) means the value is unchanged from the original value, we only highlight the values that changed.}
\label{tab:table-explanation}
\end{table}

We presented counterfactuals in a table, such as in Table~\ref{tab:table-explanation}. We show the details of a person in column \textit{Original Value} and the prediction that their income is lower than $\$50,000$. When we change the value of feature \textit{Occupation} as in columns \textit{Alternative 1} and \textit{Alternative 2}, the confidence score changes but the prediction is still lower than $\$50,000$. From this table, we can find the correlation between the \textit{Occupation} and the confidence score; the occupation \textit{Service} gives the prediction with the highest confidence score among all three occupations.

\subsection{Visualisation-Based Counterfactual Explanation}
\begin{figure}
	\centering
	\includegraphics[width=0.9\linewidth]{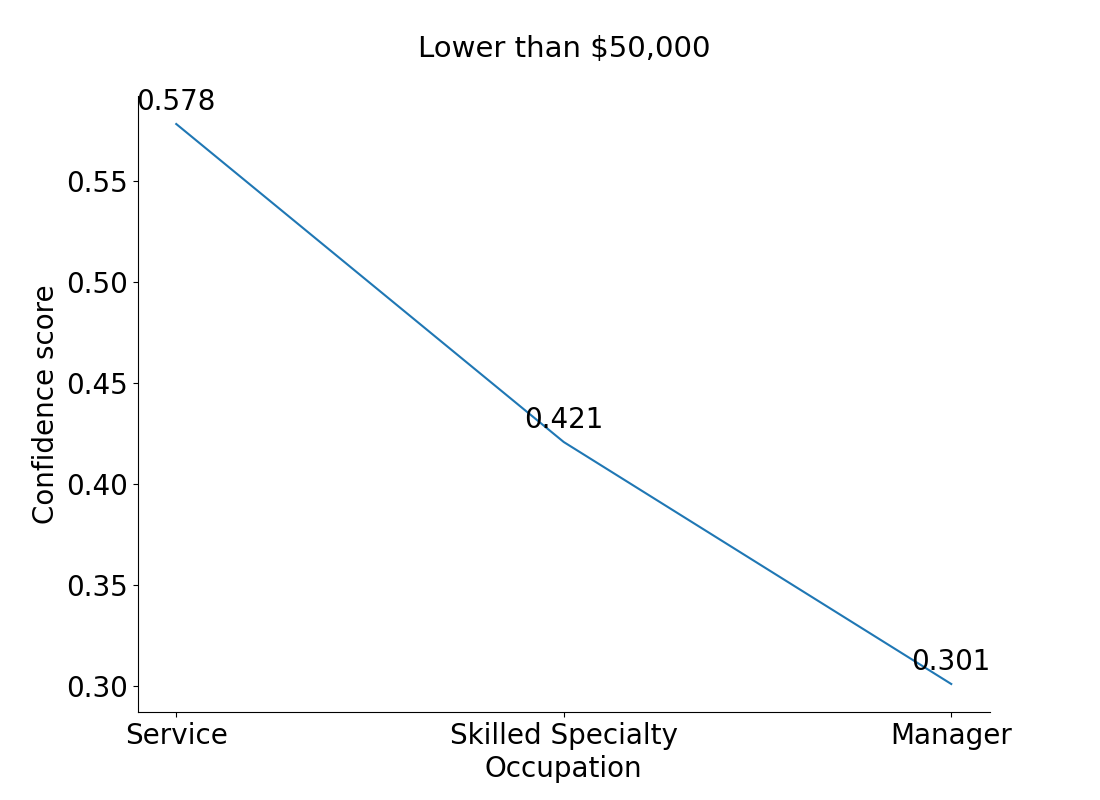}
	\caption{Counterfactual visualisation: Categorical variable}
	\label{fig:graph-explanation-discrete}
\end{figure}

\begin{figure}
    \centering
    \includegraphics[width=0.9\linewidth]{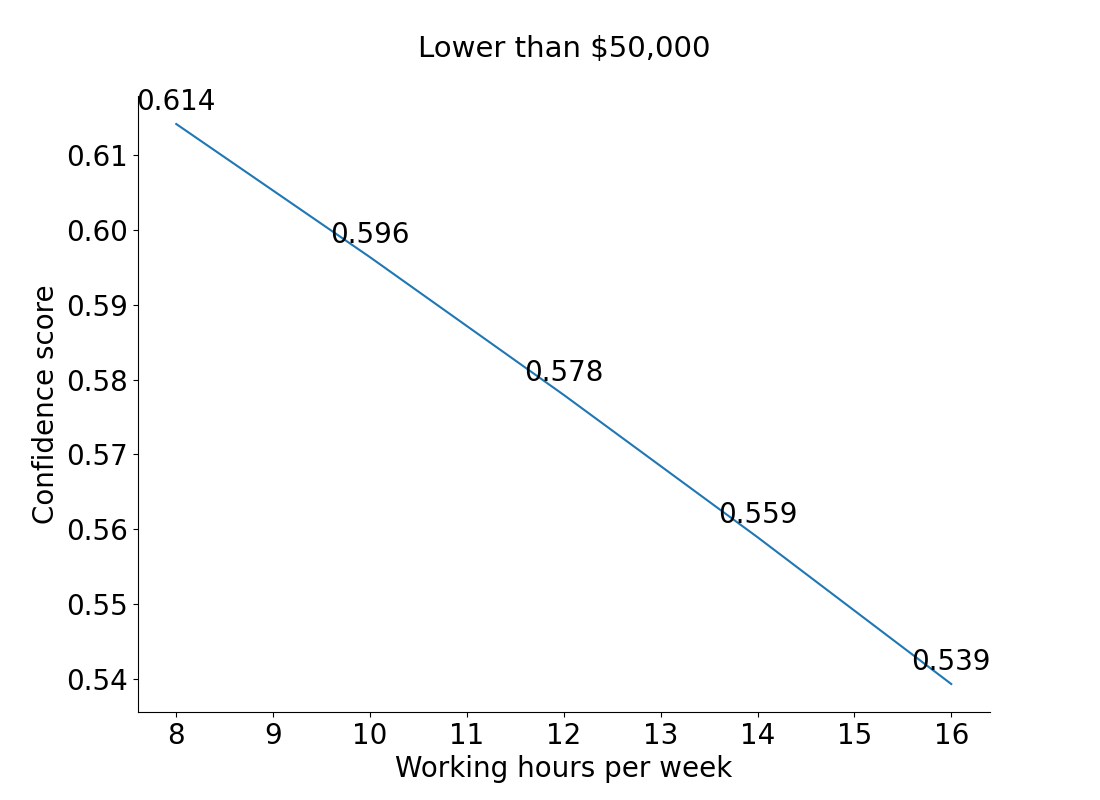}
	\caption{Counterfactual visualisation: Continuous variable}
	\label{fig:graph-explanation-continuous}
\end{figure}

In this section, we propose a method for visualising the counterfactual space of a model and how this affects the model's confidence as shown in Figure~\ref{fig:graph-explanation-discrete} and~\ref{fig:graph-explanation-continuous}. The idea is to visualise how varying a single feature affects the model's confidence, relative to the factual input $x$. For example, Figure~\ref{fig:graph-explanation-discrete} shows the visualisation based on Table~\ref{tab:table-explanation} in the income prediction task. Here we can see the prediction reaches maximum confidence score at \textit{Service} occupation. The title of this graph shows the output prediction \textit{Lower than $\$50,000$} and the feature name \textit{Occupation} which we used to change the values.

\begin{table*}[t!]
	\centering \small
	\begin{tabular}{c p{0.2\textwidth}  p{0.3\textwidth}  p{0.3\textwidth}}
		\toprule
		& \textbf{Control (C)} & \textbf{Treatment - Example-Based (E)} & \textbf{Treatment - Visualisation-Based (V)}\\
		\midrule
		Phase 1 & \multicolumn{3}{c}{
			Participants are given plain language statement, consent form and demographic questions (age, gender)
		}\\
  
		Phase 2 & \multicolumn{3}{c}{Participants are provided with}\\
		 & Input instances & Input instances & Input instances\\
		& AI model's prediction class & AI model's prediction class & AI model's prediction class\\
		& & Counterfactual examples & Counterfactual visualisation \\
		Phase 3 & \multicolumn{1}{c}{Nothing} & \multicolumn{2}{c}{10-point Likert \textit{Explanation Satisfaction Scale}} \\
		Phase 4 & \multicolumn{3}{c}{10-point Likert \textit{Trust Scale}} \\
		\bottomrule
	\end{tabular}
	\caption{Summary of participants' tasks in our three experimental conditions}
	\label{tab:task-groups}
\end{table*}

This visualisation technique is based on the idea of \textbf{Individual Conditional Expectation (ICE)}~\cite{Goldstein15}. ICE is often used to show the effect of a feature value on the predicted probability of an instance. In our study, we show how changing a feature value can change the \textit{confidence score} instead of changing the predicted probability as in the original ICE. There are two types of variables in the dataset: (1) categorical variable, and (2) continuous variable. So we define the ICE for confidence score of a single feature $x_i$ of instance $x$ such that $F(x_i) = U(x_i)$ for all $x_i$, where:
\begin{itemize}
    \item $x_i \in D$ if $x_i$ is a categorical value and $D$ is the categorical set
    \item $x_i \in [c_{\min}, c_{\min}+t, \ldots, c_{\max}]$ if $x_i$ is a continuous value; $c_{\min}$ and $c_{\max}$ are the minimum and maximum values of a continuous range and $t$ is a fixed increment.
\end{itemize}

If we use only a 2-dimensional graph, we can visualisation the changes of only one feature, whereas counterfactual examples can explain how changing multiple features simultaneously affect the confidence. However, visualising the counterfactual space allows us to easily identify the lowest and highest confidence values for categorical values and the trend of continuous values.

%% file: 4-experiment.tex
\section{Human-Subject Experiments}

Our user experiments test the following hypotheses.
\begin{itemize}
    \item \textbf{Hypothesis 1a/b (H1a/b)}: \textbf{Example-based}/\textbf{Visualisation-based} counterfactual explanations help users better \textbf{understand} the AI model than when they are not given explanations.
    \item \textbf{Hypothesis 2a/b (H2a/b)}: \textbf{Example-based}/\textbf{Visualisation-based} counterfactual explanations help users better \textbf{trust} the AI model than when they are not given explanations.
\end{itemize}

It is necessary to test against the baseline of no explanation because providing explanations is not always useful compared to not providing any explanations~\cite{LaiT19Human,Bansal-complementary21}. We then evaluate the difference between example-based counterfactual explanations and visualisation-based counterfactual explanations based on the following hypotheses.

\liz{Suggest this is a place to add something like... ``While testing against a baseline of \emph{No explanation'} may seem uninteresting, we regarded it as useful to include the check, as it is not always the case that more information helps the user \cite{XXX}.'' - where ideally you would have a ref dealing with trust and one with understanding...but even some ref illustrating the general issue should suffice}

\begin{itemize}
    \item \textbf{Hypothesis 3a/b/c (H3a/b/c)}: \textit{Visualisation-based} counterfactual explanations help users better \textbf{understand/trust/be satisfied with} the AI model than \textit{example-based} counterfactual explanations.
\end{itemize}

To evaluate {\bf understanding}, i.e., H1a, H1b and H3a, we use \textit{task prediction}~\cite[p11]{Hoffman-metrics-xai}. Participants are given some instances and their task is to decide for which instance the AI model will predict a higher confidence score. Thus, task prediction helps evaluate the user's mental model about their understanding in model confidence. 

To evaluate {\bf trust}, i.e., H2a, H2b and H3b, we use the 10-point Likert \textit{Trust Scale} from~\cite[p49]{Hoffman-metrics-xai}. For {\bf satisfaction}, i.e., H3c, we use the 10-point Likert \textit{Explanation Satisfaction Scale} from~\cite[p39]{Hoffman-metrics-xai}.

\subsection{Experimental Design}

\paragraph{Dataset} We ran the experiment on two different domains from two different datasets, which are \textit{income prediction domain} and \textit{HR domain}. Both datasets are selected so that experiments can be conducted on general participants with no requirement of particular expertise. The data used for the income prediction task is the Adult Dataset published in UCI Machine Learning Repository~\cite{Dua-2019} that includes 32561 instances and 14 features. This dataset classifies a person's income into two classes (below or above \$50K) based on personal information such as marital status, age, and education. In the second domain, we use the IBM HR Analytics Employee Attrition Performance dataset published in Kaggle~\cite{pavansubhash_2017}, which includes 1470 instances and 34 features. This dataset classifies employee attrition as yes or no based on some demographic information (job role, daily rate, age, etc.). We selected the seven most important features for both datasets by applying the Gradient Boosting Classification model over all data.

\paragraph{Model Implementation} In our experiments, we use logistic regression to calculate the probability of a class, so $P(x) = \frac{1}{1+e^{-y}}$ where $y=wx$ is a linear function of point $x$. We chose logistic regression because of its simplicity so that we can easily define the confidence score. Moreover, although logistic regression models are considered intrinsically interpretable models \cite{molnar2019}, it is still challenging to reason about their behaviour when we want to have a lower (or higher) confidence score. In future work, our studies can be extended to using counterfactual tools for more complex models, such as CLUE \cite{Antoran21explainUncertainty}.

We choose \textit{margin of confidence}, which is the difference between the first and the second highest probabilities~\cite[p93]{monarch2021human} as the formula of confidence score $U(x)$. The higher the difference between two class probabilities, the more confident the prediction is in the highest probability class.

\begin{table}[t]
	\centering \small
	\setlength{\tabcolsep}{2pt} 
	\begin{tabular}{lccc}
		\toprule
		\textbf{Attribute} & \textbf{Employee 1} & \textbf{Employee 2} & \textbf{Employee 3} \\
		\midrule
		Marital status & Married & Married & Married \\
		Years of education & 15 & 15 & 15\\
		Occupation & \textbf{Service} & \textbf{Manager} & \textbf{Skilled}\\ 
        & & & \textbf{Specialty} \\
		Age & 25 & 25 & 25\\
		Any capital gains & No & No & No\\
		Working hours & 30 & 30 & 30\\
		per week & & \\
		Education & Bachelors &  Bachelors & Bachelors\\
		\midrule
		\textbf{AI model prediction} & \multicolumn{3}{c}{\textbf{Lower than }\bm{$\$50,000$}} \\
		\bottomrule
	\end{tabular}
	\caption{Example input instances provided in the question. The question is: ``For which employee the AI model predicts with the highest confidence score?"}
	\label{tab:example-question}
\end{table}

\paragraph{Procedure} Before conducting the experiments, we received ethics approval from our institution. We recruited participants on Amazon Mechanical Turk (Amazon MTurk), a popular crowd-sourcing platform for human-subject experiments~\cite{buhrmester2016amazon}. The experiment was designed as a Qualtrics survey\footnote{\url{https://www.qualtrics.com/}} and participants can navigate to the survey through the Amazon MTurk interface. We allowed participants 30 minutes to finish the experiment and paid each participant a minimum of USD \$7 for their time, plus a maximum of up to USD \$2 depending on their final performance.

We use a between-subject design such that participants were randomly assigned into one of three groups: (1) \textit{Control (C)}; (2) \textit{Treatment with Example-Based Explanation (E)}; or (3) \textit{Treatment with Visualisation-Based Explanation (V)}. For each group, there are four phases that are described in Table~\ref{tab:task-groups}. The difference between the control group and the treatment group is that in the control group, participants were not given any explanations. In the task prediction (phase 2), participants in the control group were only shown input values along with the AI model prediction class as in Table~\ref{tab:example-question}. 
In the treatment group, participants were provided with either example-based explanations (e.g. Table~\ref{tab:table-explanation}) or visualisation-based explanations (e.g. Figure ~\ref{fig:graph-explanation-discrete}). The participants each received the same 10 questions. For each question, they were asked to select an input instance out of \textit{3 instances} that the AI model would predict with the highest confidence score. A question can have either one or two explanations depending on the number of modified attributes in the question. For instance, the question in Table~\ref{tab:example-question} changes only one attribute \textit{Occupation} so participants were given a single explanation in treatment conditions. An explanation can either present a \textit{categorical variable} (e.g. Figure~\ref{fig:graph-explanation-discrete}) or a \textit{continuous variable} (e.g. Figure~\ref{fig:graph-explanation-continuous}).

We scored each participant using: 1 for a correct answer, -2 for a wrong answer and 0 for selecting ``I don't have enough information to decide". To imitate high-stake domains, the loss for a wrong choice is higher than the reward for a correct choice~\cite[p2433]{Bansal19}. They are also asked to briefly explain why they choose that option in a text box, which is analysed later in the qualitative analysis. The final compensation was calculated based on the final score --- a score of $0$ or less than $0$ received \$7 USD and no bonus. A score greater than $0$ received a bonus of \$0.2 for each additional score.

\paragraph{Participants} We recruited a total of 180 participants for two domains, that is 90 participants for each domain from Amazon MTurk. Then 90 participants were evenly randomly allocated into three groups (30 participants in each group). All participants were from the United States. We only recruited Masters workers, who achieved a high degree of success in their performance across a large number of Requesters~\footnote{\url{https://www.mturk.com/worker/help}}. For the \textit{income prediction domain}, 41 participants were women, 1 was self-specified as non-binary, 48 were men. Between them, 4 participants were between Age 18 and 29, 34 between Age 30 and 39, 27 between Age 40 and 49, 25 over Age 50. For the \textit{HR domain}, 43 participants were women, 47 were men. Age wise, 4 participant was between Age 18 and 19, 37 between Age 30 and 39, 26 between Age 40 and 49, 23 over Age 50.

We performed \textit{power analysis} for two independent sample t-test to determine the needed sample sizes. We calculate the Cohen's d between control and treatment group and obtain the effect size of $0.7$ and $0.67$ in income and HR domain. Using power of $0.8$ and significant alpha of $0.05$, we get sample sizes of $26$ and $29$ in the two domains. Thus, we determine the sample size needed for a group is $30$ and the total number of samples needed is $90$ for one domain.

\subsection{Results: Summary of Both Domains}

\begin{figure}[t!]
    \begin{subfigure}[b]{0.33\linewidth}
		\centering
		\includegraphics[width=\linewidth]{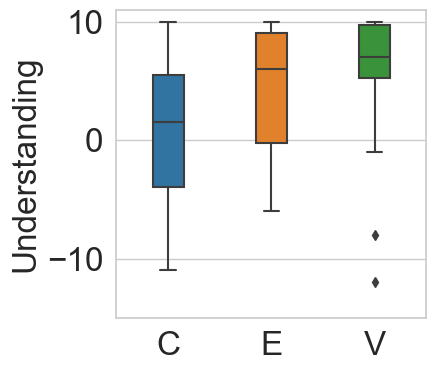}
		\label{fig:income-final-score}
    	\end{subfigure}
    	\begin{subfigure}[b]{0.33\linewidth}
    	\centering
     \includegraphics[width=\linewidth]{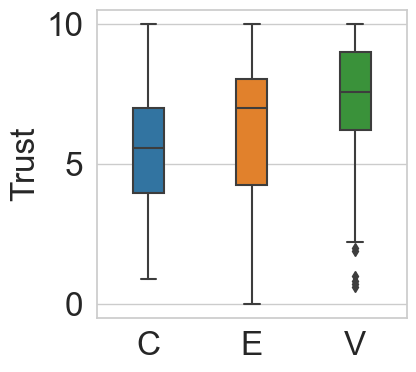}
     \label{fig:income-trust-stack}
    	\end{subfigure}
     \begin{subfigure}[b]{0.32\linewidth}
    	\centering
     \includegraphics[width=\linewidth]{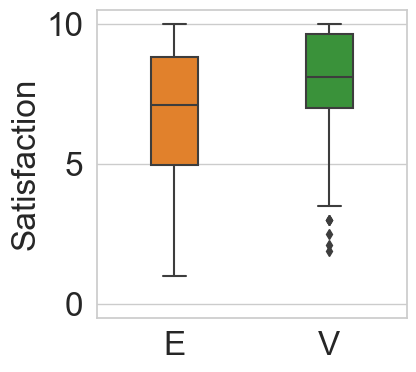}
    	\label{fig:income-explanation-stack}
    \end{subfigure}
    \caption{Domain 1 (Income)}
    \label{fig:income-results}  
\end{figure}

\begin{figure}[t!]
    \begin{subfigure}[b]{0.33\columnwidth}
		\centering
		\includegraphics[width=\linewidth]{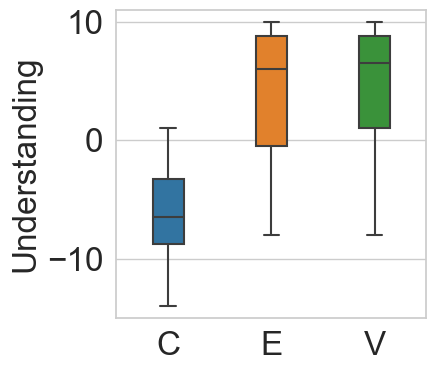} 
		\label{fig:hr-final-score}
	\end{subfigure}
	\begin{subfigure}[b]{0.33\columnwidth}
		\centering
		\includegraphics[width=\linewidth]{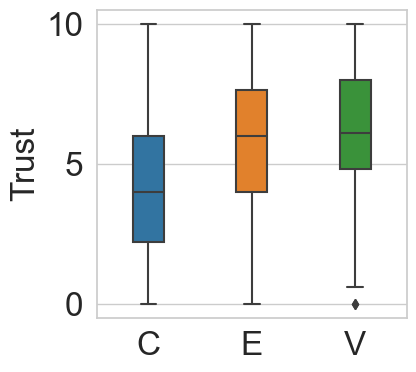}
		\label{fig:hr-trust-stack}
	\end{subfigure}
 	\begin{subfigure}[b]{0.32\columnwidth}
		\centering
		\includegraphics[width=\linewidth]{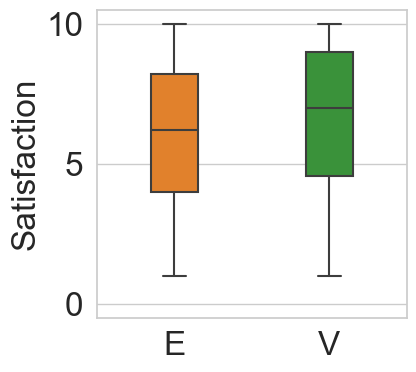} 
		\label{fig:hr-explanation-stack}
	\end{subfigure}
	\caption{Domain 2 (HR)}
	\label{fig:hr-results}
\end{figure}

In this section, we present the results from our experiment for two domains that used the income and HR datasets. We tested for data normality by using the Shapiro-Wilks test and found that our data was not normally distributed. Therefore, we applied the Mann–Whitney U test, which is a non-parametric test equivalent to the independent samples t-test to perform pairwise comparisons between two groups. Table~\ref{tab:hypotheses} summarises our results of testing the seven hypotheses. Figure~\ref{fig:income-results} and~\ref{fig:hr-results} show the results of the two studies. 

\begin{table}[t]
	\centering \small
	\setlength{\tabcolsep}{3pt} 
\begin{tabular}{lccccccc}
    \toprule
  &  \multicolumn{3}{c}{Understanding} & \multicolumn{3}{c}{Trust} & Satisfaction \\
    \cmidrule(lr){2-4} \cmidrule(lr){5-7} \cmidrule(lr){8-8}
   & \textbf{H1a} & \textbf{H1b} & \textbf{H3a} & \textbf{H2a} & \textbf{H2b} & \textbf{H3b} & \textbf{H3c}\\
  & E & V & E vs V & E & V & E vs V & E vs V \\
  \midrule
\textbf{Domain 1} & $\checkmark$ & $\checkmark$ & $\times$ & $\checkmark$ & $\checkmark$ & $\checkmark$ & $\checkmark$ \\
\textbf{Domain 2} & $\checkmark$ & $\checkmark$ & $\times$ & $\checkmark$ & $\checkmark$ & $\times$ & $\times$ \\
\bottomrule
\end{tabular}
\caption{Summary of hypothesis tests in two domains.  $\checkmark$ represents the hypothesis is supported, $\times$ represents the hypothesis is rejected.}
\label{tab:hypotheses}
\end{table}

\begin{table*}[t]
	\centering \small
	\setlength{\tabcolsep}{2pt} 
	\renewcommand{\arraystretch}{1.2} 
	\begin{tabular}{p{0.25\textwidth}p{0.74\textwidth}}
		\toprule
		\textbf{Code} & \textbf{Definition} \\
		\midrule
	    \textbf{W-Reversed category} (CAT) & The participant selected the instance that has a lowest confidence score instead of  a highest confidence score among all instances \\
		\textbf{W-Linear assumption} (CAT) & Assumed the correlation between confidence score and attribute values was linear when it was not (e.g. the feature was categorical) \\
		\textbf{W-Small differences} (CAT \& CON) & Selected a wrong answer due to small differences in the explanation and/or the question \\
		\textbf{W-Reversed correlation} (CON) & Reversed the trend of the explanation of a continuous variable \\
		\textbf{W-Case-based} (CON) & Used case-based reasoning when the correlation was linear \\
		\midrule
		\textbf{D-No correlation} (CAT \& CON) & Could not find the trend of the confidence score \\
		\textbf{D-Different attribute values} (CAT \& CON) & Argued that the values of instances in the explanations are not the same as values in the question \\
		\textbf{D-Outside range} (CON) & The modified values in the question are beyond the lowest and highest values in the explanation \\
		\midrule
		\textbf{C-Correlation-based} (CON) & Found the correlation in the explanation \\
		\textbf{C-Case-based} (CAT) & Got the correct answer based on examples in the explanation without mentioning about the correlation \\
		\bottomrule
	\end{tabular}
	\caption{The codebook for participants' responses to evaluate how they understand the provided explanations. \textit{CAT}, \textit{CON} mean the code is applied for categorical variables and continuous variables, respectively. \textit{W} corresponds to wrong answers. \textit{D} corresponds to the ``do not have enough information to decide". \textit{C} corresponds to correct answers.}
	\label{tab:codes}
\end{table*}

\textbf{The results show that counterfactual explanations of confidence scores help users understand and trust the AI model more than those who were not given counterfactual explanations}. We conclude that \textbf{H1a, H1b, H2a} and \textbf{H2b} are supported in both studies ($p<0.005$, $r>0.41$).

\textbf{There is no statistically significant difference in improving users' understanding between example-based explanations and visualisation-based explanations} --- \textbf{H3a} is rejected. In domain 1, the difference in the task prediction between the two treatment groups is larger than that in domain 2. Specifically, effect size in domain 1 is $r=0.23$ ($p=0.13$) and in domain 2 is $r=0.03$ ($p=0.86$).

\textbf{There are some discrepancies between domain 1 and 2 when comparing example-based and visualisation-based explanations in terms of trust and satisfaction}. In the first domain, \textbf{H3b} ($p<0.001, r=0.26$) and \textbf{H3c} ($p<0.001, r=0.28$) are supported. However, in domain 2, \textbf{H3b} ($p=0.1>0.05$) and \textbf{H3c} ($p=0.06>0.05$) are both rejected. We envision the discrepancies of H3b and H3c may be because prior knowledge of participants could affect them doing the tasks in two different domains. Future work could test this idea further. 

As observing no statistically significant difference between example-based and visualisation-based explanations, we then use qualitative analysis to find the limits of both designs and suggest directions to design effective explanations.

\subsection{Qualitative Analysis}

\begin{table*}[h]
	\centering \small
	\setlength{\tabcolsep}{2pt} 
\begin{tabular}{lllr @{\hspace{\tabcolsep}} r @{\hspace{4\tabcolsep}} r @{\hspace{\tabcolsep}} r @{\hspace{4\tabcolsep}} r @{\hspace{\tabcolsep}} r @{\hspace{4\tabcolsep}} r @{\hspace{\tabcolsep}} r}
  & & & \multicolumn{4}{c}{Income} &  \multicolumn{4}{c}{HR} \\
    \cmidrule(lr){4-7} \cmidrule(lr){8-11}
  & & & \multicolumn{2}{c}{E} & \multicolumn{2}{c}{V} & \multicolumn{2}{c}{E} & \multicolumn{2}{c}{V} \\
  \midrule

\multirow{6}{*}{Wrong Answer} 
    & \multirow{3}{*}{Categorical Variables} & W-Linear assumption & 0 & (0\%) & 0 & (0\%) & 39 & (95\%) & 21 & (78\%) \\ 
& & W-Small difference & 5 & (28\%) & 0 & (0\%) & 2 & (5\%) & 2 & (7\%) \\
& & W-Reversed category & 13 & (72\%) &	11 & (100\%) & 0 & (0\%) & 4 & (15\%) \\
    \cmidrule(lr){2-11}
    & \multirow{3}{*}{Continuous Variables} & W-Case-based & 1 & (4\%) & 0 & (0\%) &	7 & (64\%) & 0 & (0\%) \\
& & W-Small difference & 0 & (0\%) & 2 & (18\%) & 0 & (0\%) & 2 & (12\%) \\
& & W-Reversed correlation & 23 & (96\%) & 9 & (82\%) & 4 & (36\%) & 14 & (88\%) \\
\midrule

\multirow{4}{*}{Not Enough Information} & \multirow{2}{*}{Categorical Variables} & D-Different attribute values & 6 & (100\%) & 0 & & 8 & (80\%) & 0 & \\
& & D-No correlation & 0 & (0\%) & 0 & & 2 & (20\%) & 0 & \\
\cmidrule(lr){2-11}
& \multirow{3}{*}{Continuous Variables} & D-Outside range & 1 & (6\%) & 9 & (64\%) & 2 & (13\%) & 6 & (46\%) \\
& & D-Different attribute values & 4 & (24\%) & 0 & (0\%) & 3 & (19\%) & 0 & (0\%) \\
& & D-No correlation & 12 & (70\%) & 5 & (36\%) & 11 & (68\%) & 7 & (54\%) \\
\midrule

\multirow{2}{*}{Correct Answer} & \multirow{2}{*}{Categorical Variables} & C-Correlation-based & 17 & (11\%) & 20 & (11\%) & 18 &(10\%) & 0 & (0\%) \\
& & C-Case-based & 133 & (89\%) & 159 & (89\%) & 157 & (90\%) & 186 & (100\%) \\
\cmidrule(lr){2-11}
 & \multirow{2}{*}{Continuous Variables} & C-Correlation-based &	97 & (98\%) & 118 & (100\%) & 81 & (98\%) & 99 & (100\%) \\
& & C-Case-based & 2 & (2\%) & 0 & (0\%) & 2 & (2\%) & 0 & (0\%) \\
\bottomrule
\end{tabular}
\caption{Frequencies and Percentages of Codes for Explanations}
\label{tab:categorical-continuous}
\end{table*}

We perform the thematic analysis~\cite{Braun2006-bm} from the text written by participants after each mutiple-choice question to know why they selected an option. The text is a response to ``Can you please explain why you selected this option?". We followed ~\citet{Nowell2017-gv} who gave a step-by-step approach for doing trustworthy thematic analysis. Three authors were involved in the qualitative analysis. The first author identified and documented the themes and the codes. Through multiple discussion meetings, two other authors critically analysed the codes and verified them. Finally, we decided on the final codes as in Table~\ref{tab:codes}.

Every participant did the same 10 questions so we have 30 (participants) $\times$ 10 (questions) is 300 (texts) for a condition. Given that we have two treatment conditions and two datasets, we analysed a total of 1,200 texts and each text is assigned to one code or more than one code depending on the number of explanations in that response. 
Each code is classified as one of: (1) a correct answer (C); (2) a wrong answer (W); or (3) ``not enough information" (D). The final analysis includes 1,112 texts after removing 88 texts due to poor quality. We found the following observations, which suggest future improvements.

\textbf{Use text labels instead of numbers to present categorical variables.}. A categorical variable can be shown in numbers or text labels. In Table~\ref{tab:categorical-continuous}, the majority of wrong codes in HR domain is \textit{W-Linear assumption} (78\% and 95\%) because most explanations using categorical variables are written in numbers. There was no \textit{linear assumption} codes in the income dataset since all explanations used text labels.

\textbf{When the labels of categorical features indicate ordinal data, visualise counterfactuals help to reduce the error ``linear assumption'', making it easier for people to interpret the highest or lowest values}. According to Table~\ref{tab:categorical-continuous} (HR dataset), $95\%$ (39) of wrong responses happened due to \textit{linear assumption} in the example-based condition; however, we found only $78\%$ (21) of \textit{linear assumption} in the visualisation-based condition. For instance, in a question where the job level is a categorical variable and is not correlated with the confidence score, a participant in the example-based condition mentioned: \textit{``Those with a higher job level had a higher confidence rating"}. In contrast, another participant in the visualisation-based condition could identify the highest confidence value at job level 2 without mentioning about the linear trend: \textit{``The AI predicted job level 2 has the highest chance of staying"}.

\textbf{It is hard for people to interpret the example-based explanations when the differences between counterfactual outputs and categorical attributes are minimal}. According to Table~\ref{tab:categorical-continuous} (wrong answer, income dataset, categorical variables), we observe $28\%$ (5) of \textit{W-Small difference} codes in the example-based condition. For example, in a question where \textit{Manager} occupation has the highest confidence score, some participants mistakenly selected \textit{Skilled Specialty} as the highest even though this occupation is the second highest. In this case, the difference in confidence values between \textit{Manager} and \textit{Skilled Specialty} is only 2\% (93\% and 91\%). This small difference made 5 participants chose a wrong answer in the example-based condition.

\textbf{Using visualisation-based explanations is easier to understand correlations; however, many participants were not willing to extrapolate the correlation beyond the lowest and highest values}. In Table~\ref{tab:categorical-continuous} (Not Enough Information), we have fewer codes of \textit{D-No correlation} in visualisation-based explanations. However, we record a higher number of codes of \textit{D-Outside range} in this visualisation condition. This issue suggests that we should not expect participants to extrapolate the correlation, and all counterfactual points should be shown in the explanations.

\textbf{Regardless of variables, if the counterfactual examples in the example-based explanations are not the same as the values in the question, many participants argued that they do not have enough information to decide} (\textit{D-Different attribute values}). For example, a participant said: \textit{``Because the position is different, lab tech versus sale rep, I feel that even though the AI chose the one with the highest confidence as the one with the lowest daily rate, I am not sure if the job description would change that confidence level"}. In this question, we provided the example-based explanation for \textit{Sales Representative} job, but the question shows instances of \textit{Lab Tech} job. Even though the daily rate increases linearly in all cases, some participants did not feel confident to apply this observation when we change the instance values in the question. They applied \textit{case-based reasoning} when interpreting the example-based explanation of a linear model rather than interpreting the linear correlation in this explanation. That is, they found the closest example in the counterfactual explanation presented, and compared that example with the question. Similarly, we found an overall 8 codes of \textbf{W-Case-based} where participants applied case-based reasoning to do the task with example-based explanations of continuous values. A participant wrote: \textit{``It really is a tough call but I chose employee 1 because the 400 range has the highest percentage of leaving"}. In this example, the participant saw that the daily rate of 400 has the highest confidence of leaving, therefore, they selected the value that is close to 400 rather than interpreting the linear correlation between the daily rate and the confidence score (lower daily rate indicates higher confidence of leaving). Specifically, the question has three daily rate options of 200, 201 and 247, they eventually selected 247 as the final answer, arguing that 247 is closest to 400 in the example-based explanation. In general, \textbf{it is clear that participants in the example-based condition used a `case-based reasoning' approach to understanding the model}. This led participants to overlook the linear trend between the confidence score and the feature values. This finding suggests that we should be careful when using example-based explanations to interpret continuous variables for models, except for cases when the underlying model is itself a case-based model. Using graphs to visualise continuous variables can mitigate this issue.

%% file: 6-conclusion.tex
\section{Conclusion}
This paper proposes two approaches for counterfactual explanation of model confidence: (1) example-based counterfactuals; and (2) visualisation-based counterfactuals. Through a human-subject study, we show that the counterfactual explanation of model confidence helped users improve their understanding and trust in the AI model. Furthermore, the qualitative analysis suggests directions of designing better counterfactual explanations. In the future, we plan to perform more extensive user studies to evaluate whether we can improve decision making using such explainability techniques.